\documentclass[11pt,a4paper]{article}
\usepackage[hyperref]{naaclhlt2019}
\usepackage{times}
\usepackage{latexsym}
\usepackage{url}
\usepackage{graphicx}
\usepackage{multirow}
\usepackage{multicol}
\usepackage{amsmath}
\usepackage{amsthm}
\usepackage{amssymb,color}
\usepackage{bm}
\aclfinalcopy 

\title{Dirichlet Variational Autoencoder for Text Modeling}
\author{Yijun Xiao \\
   University of California, Santa Barbara\\
  {\tt yijunxiao@cs.ucsb.edu} \\
  \And Tiancheng Zhao \\
  Language Technologies Institute \\
Carnegie Mellon University \\
{\tt tianchez@cs.cmu.edu}\\
  \AND William Yang Wang\\
   University of California, Santa Barbara\\
  {\tt william@cs.ucsb.edu}\\
}

\date{}

\newcommand{\vect}[1]{\mathbf{#1}}
\newcommand{\matr}[1]{\mathbf{#1}}

\newcommand{\vh}[0]{\vect{h}}

\newcommand{\vt}[0]{\vect{t}}

\newcommand{\vx}[0]{\vect{x}}

\newcommand{\vz}[0]{\vect{z}}
\newcommand{\vzero}[0]{\vect{0}}
\newcommand{\valpha}{\boldsymbol{\alpha}}
\newcommand{\vtheta}{\boldsymbol{\theta}}
\newcommand{\vphi}{\boldsymbol{\phi}}
\newcommand{\vmu}{\boldsymbol{\mu}}
\newcommand{\vsigma}{\boldsymbol{\sigma}}

\newcommand{\vlb}{\mathcal{L}}

\newcommand{\E}{\mathbb{E}}

\newcommand{\N}{\mathcal{N}}

\newcommand{\mX}[0]{\matr{X}}

\newcommand{\eye}{\matr{I}}
\newcommand\dir{\text{Dir}}
\newcommand\kl{\textsc{kl}}

\newcommand\bow{\textsc{bow}}
\newcommand\nll{\textsc{nll}}
\newcommand\ppl{\textsc{ppl}}
\newcommand\kld{\text{D}_{\textsc{kl}}}
\newcommand\vae{\textsc{vae}}
\newcommand\cvae{\textsc{cvae}}
\newcommand{\lstm}{\textsc{lstm}}
\newcommand\lm{\textsc{lm}}
\newcommand\elbo{\textsc{elbo}}

\newcommand{\cnn}{\textsc{cnn}}
\newcommand{\mlp}{\textsc{mlp}}
\newcommand{\lda}{\textsc{lda}}
\newcommand{\svm}{\textsc{svm}}
\newcommand{\lr}{\textsc{lr}}
\newcommand{\ptb}{\textsc{ptb}}
\newcommand{\bc}{\textsc{bc}}
\newcommand{\yelp}{\textsc{yelp}}
\newcommand{\yahoo}{\textsc{yahoo}}
\newcommand\diag{\text{diag}}

\begin{document}
\maketitle
\begin{abstract}
We introduce an improved variational autoencoder (\vae) for text modeling with topic information explicitly modeled as a Dirichlet latent variable. By providing the proposed model topic awareness, it is more superior at reconstructing input texts. Furthermore, due to the inherent interactions between the newly introduced Dirichlet variable and the conventional multivariate Gaussian variable, the model is less prone to \kl{} divergence vanishing.
We derive the variational lower bound for the new model and conduct experiments on four different data sets. The results show that the proposed model is superior at text reconstruction across the latent space and classifications on learned representations have higher test accuracies.
\end{abstract}

\section{Introduction}
The variational autoencoder (\vae) \cite{kingma2013auto,rezende2015variational} is a generative model that can be seen as a regularized version of the standard autoencoder. \vae{} models are able to learn meaningful representations from the data in an unsupervised fashion. A typical \vae{} samples latent representations of the data from a multivariate Gaussian (continuous) or a multi-way categorical (discrete), and parameterizes the generator with a deep neural network. With the variational objective, the latent representations learned with \vae{} are encouraged to be smoothly transitioned instead of isolated in the latent space. Using the reparameterization trick \cite{kingma2013auto,rezende2015variational}, the inference network could be trained with back-propagation. \vae{} has been successfully applied to train generative models on image data \cite{gregor2015draw,yan2016attribute2image}. The application of \vae{} on text data are explored but less successful \cite{bowman2015generating,miao2016neural,yang2017improved,zhao2017learning}. One key issue when training \vae{} on text data is a phenomenon called Kullback-Leibler (\kl{}) divergence vanishing. With a long short-term memory (\lstm{}) \cite{hochreiter1997long} decoder, the \kl{} divergence term in the training objective quickly collapses to zero. This phenomenon leads to an obsolete encoder which produces random representations regardless of the inputs and a well-trained decoder which is essentially a language model.

Many efforts were made to address the issue of \kl{} divergence vanishing. 
These efforts were mostly made by trying different neural structures \cite{yang2017improved,semeniuta2017hybrid} or modifying the objective functions \cite{bowman2015generating,zhao2017learning,zhao2018unsupervised}. We propose a new \vae{} model for text that incorporates topic distributions into the framework. A Dirichlet latent variable is introduced to represent the topic distributions of the input documents and the decoder has to make use of both the multivariate Gaussian variable and the Dirichlet variable to reconstruct the documents. Intuitively, the generative part of the model is similar to how human construct a sentence: we determine the sentence topics based on our abstract thoughts; and words are selected and arranged given the topics and abstract beliefs to construct the sentence. The formulation of the model requires it to predict the topic distributions based on the multivariate Gaussian variable. The \kl{} divergence vanishing problem is hence effectively handled because topics can not be accurately predicted if latent representations do not contain enough information of the input document. We derived the variational lower bound for the proposed model and conducted experiments on datasets with and without labels. 
Our contributions are:
\begin{enumerate}
\item We present a novel variational autoencoder for text modeling with topic awareness.
\item We experiment both \vae{} and conditional variational autoencoder (\cvae{}) based on the proposed model on several datasets. The proposed model outperforms baselines with respect to reconstruction, representation learning, and random sample quality.
\item Further model analysis shows that the proposed model is able to generate convincing topic distributions for unseen test data.
\end{enumerate}

\section{Related Work}
\subsection{Variational Autoencoder}
Variational autoencoders have been widely explored for various machine learning tasks, e.g., image generation \cite{gregor2015draw,mansimov2015generating,yan2016attribute2image}, machine translation \cite{zhang2016variational}, knowledge graph reasoning \cite{zhang2017variational,chen2018variational}, conversational models \cite{serban2017hierarchical,zhao2017learning} and text style-transfer \cite{hu2017toward}. 
\vae{} can also be applied in semi-supervised learning \cite{kingma2014semi} by treating the class label as a variable and marginalizing over it whenever the label is not present. With Gumbel-softmax \cite{jang2016categorical,maddison2016concrete}, the marginalization can be approximated with sampling over the categorical distribution. This study does not involve extensions of the proposed model in the semi-supervised setting.

Many studies observed that the \kl{} divergence term of the \vae{} objective often vanishes when trained on text data. The posterior of the latent code $q(\vz|\vx)$ is almost identical to the prior $p(\vz)$. This leads to a decoder which completely ignores the latent code $\vz$ and trained solely to minimize the perplexity of the corpus. In this case, the model collapses to a language model.

There are two main categories of solutions to mitigate the issue. The first category is to weaken the decoder which includes randomly dropping words when feeding the decoder \cite{bowman2015generating}, and using convolutional neural networks (\cnn) in the decoder \cite{yang2017improved,semeniuta2017hybrid}. This category tackles the problem mostly in the model architecture level. The second category is to modify the original variational learning objective. \kl{} divergence term annealing \cite{bowman2015generating} can be seen as gradually adding the \kl{} divergence term into the objective function. Bag-of-word (\bow) loss introduced in \cite{zhao2017learning} is an auxiliary loss which measures how well predictions of words can be made from the latent code $\vz$. By rewriting the objective, Zhao et al. \shortcite{zhao2018unsupervised} found that the original variational lower bound is anti-information and proposed to compensate the objective function with an added mutual information term. 
The model proposed in this study does not explicitly attack \kl{} divergence vanishing, but the interaction between the Dirichlet topic variable and the multivariate Gaussian variable makes it less prone to the issue. Furthermore, the proposed model is derived within the variational inference framework and agnostic to any structure choices and loss modifications mentioned above.

\subsection{Topic Modeling}
Topic models \cite{blei2012probabilistic} aim to learn unsupervised representations of text data, and have been applied in various settings from scientific literature \cite{blei2007correlated} to natural scenes \cite{fei2005bayesian}. There are recent studies focusing on learning topic models within the variational inference framework \cite{srivastava2017autoencoding}. Although our model is able to predict topic distributions for unseen data, it is not designed to learn topic distributions. It utilizes topics generated from an external or internal topic model to learn better representations of the input texts. Therefore, we do not compare with recent studies on topic modeling.

\section{Our Approach}
In this section, we introduce the model and its corresponding training objectives. The proposed model incorporates topic modeling results into training and is more robust against \kl{} divergence vanishing. For convenience and clarity, subscripts $\vtheta, \vphi$ and superscripts $i$ are dropped when there is no ambiguity.
\subsection{Background}
Consider the dataset $\mX=\{\vx^{(i)}\}_{i=1}^N$ which are i.i.d. samples from a random variable $\vx$. Assume that $\vx$ is dependent on some unobserved continuous random variable $\vz$. Then the marginal distribution of $\vx$ can be written as:
\begin{align}
p_{\vtheta}(\vx)=\int p_{\vtheta}(\vz)p_{\vtheta}(\vx|\vz)d\vz
\end{align}
In practice this integral is intractable. Variational autoencoders \cite{kingma2013auto,rezende2015variational} propose to use a recognition model $q_{\vphi}(\vz|\vx)$ to approximate the true posterior $p_{\vtheta}(\vz|\vx)$. Instead of maximizing the marginal likelihood directly, the objective becomes the variation lower bound (also called evidence lower bound (\elbo{})) of the marginal:
\begin{align}
\label{equ:vlb}
\log{p_{\vtheta}(\vx^{(i)})}\geq &\vlb(\vtheta,\vphi;\vx^{(i)}) \nonumber\\
= &\E_{q_{\vphi}(\vz|\vx^{(i)})}[\log{p_{\vtheta}(\vx^{(i)}|\vz)}] \nonumber\\
&-\kld(q_{\vphi}(\vz|\vx^{(i)})||p_{\vtheta}(\vz))
\end{align}
The family of the prior $p_{\vtheta}(\vz)$ and posterior $q_{\vphi}(\vz|\vx)$ is often chosen so that the \kl{} divergence term can be analytically calculated. $\E_{q_{\vphi}(\vz|\vx)}[\log{p_{\vtheta}(\vx|\vz)}]$ can be approximated by sampling from $q_{\vphi}(\vz|\vx)$. We could then optimize the objective with any gradient-based algorithms.

Conditional variational autoencoder (\cvae) \cite{sohn2015learning} is a straightforward extension within the \vae{} framework. If every input $\vx$ is accompanied with a discrete label $y$, we might be interested in modeling the conditional probability of $p(\vx|y)$. The variational lower bound can be easily adapted from Equation \ref{equ:vlb}:
\begin{align}
\label{equ:cvlb}
&\vlb(\vtheta,\vphi;\vx^{(i)},y^{(i)})
= \nonumber\\
&\E_{q_{\vphi}(\vz|\vx^{(i)},y^{(i)})}[\log{p_{\vtheta}(\vx^{(i)}|\vz,y^{(i)})}] \nonumber\\
&-\kld(q_{\vphi}(\vz|\vx^{(i)},y^{(i)})||p_{\vtheta}(\vz|y^{(i)}))
\end{align}

\begin{figure}[t]
\centering
\begin{minipage}[b]{0.14\textwidth}
\centering
\includegraphics[width=0.88in]{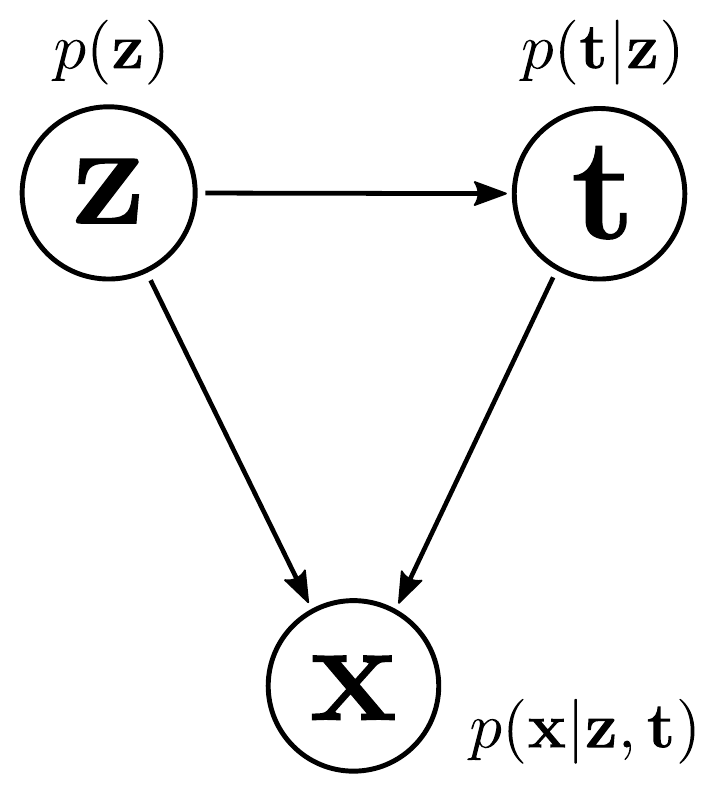}
\\
(a)
\end{minipage}
\hfill
\begin{minipage}[b]{0.14\textwidth}
\centering
\includegraphics[width=0.9in]{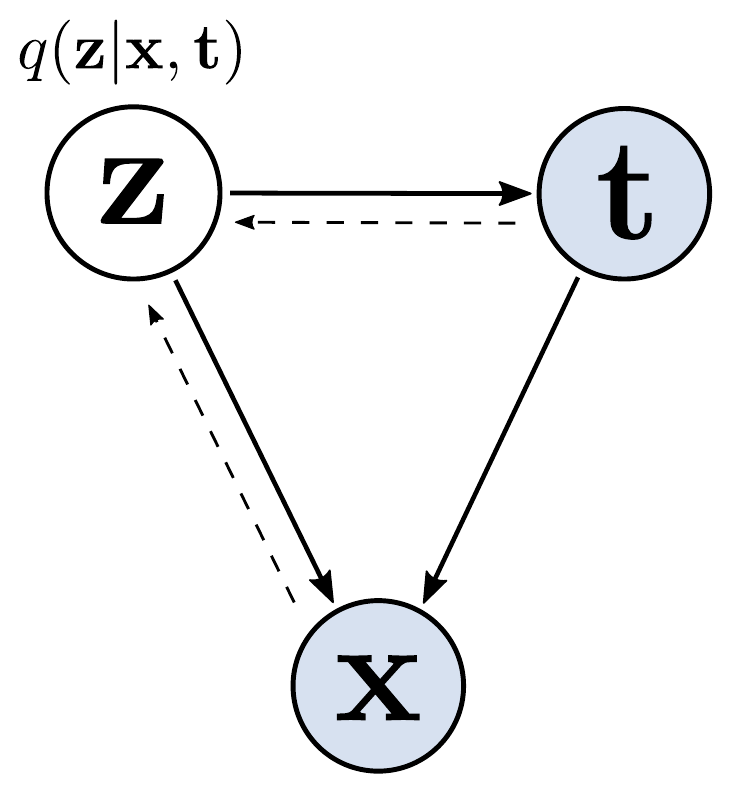}
\\
(b)
\end{minipage}
\hfill
\begin{minipage}[b]{0.14\textwidth}
\centering
\includegraphics[width=0.9in]{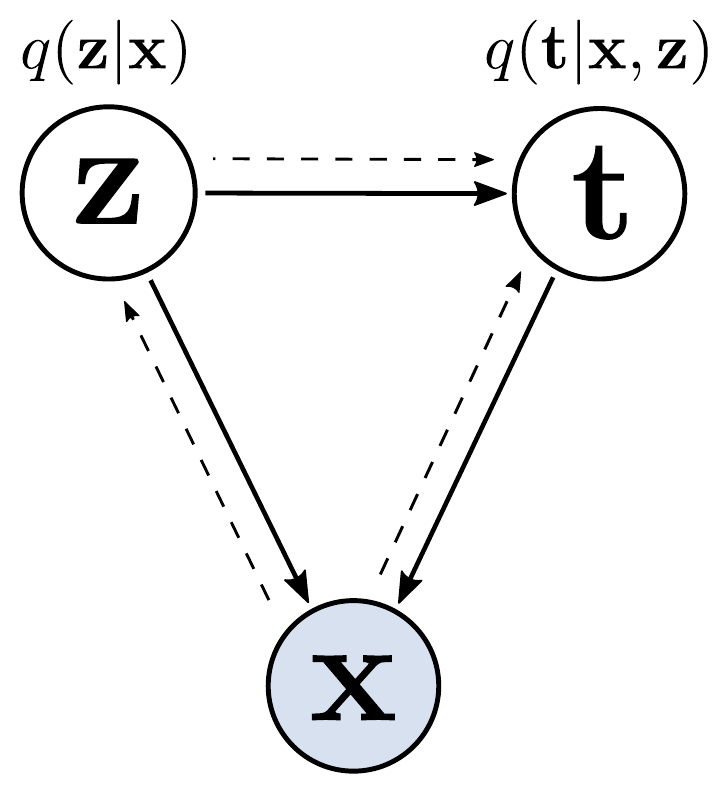}
\\
(c)
\end{minipage}
\caption{Graphical illustration of (a) the generative model (b) the joint recognition model 
and (c) the marginal recognition model. Solid lines are the shared generative model; dashed lines are the respective recognition model; shaded nodes represent observed variables.}
    \label{fig:models}
\end{figure}

\begin{figure*}\centering    
\begin{minipage}[b]{0.45\textwidth}
\centering
\includegraphics[width=2.6in]{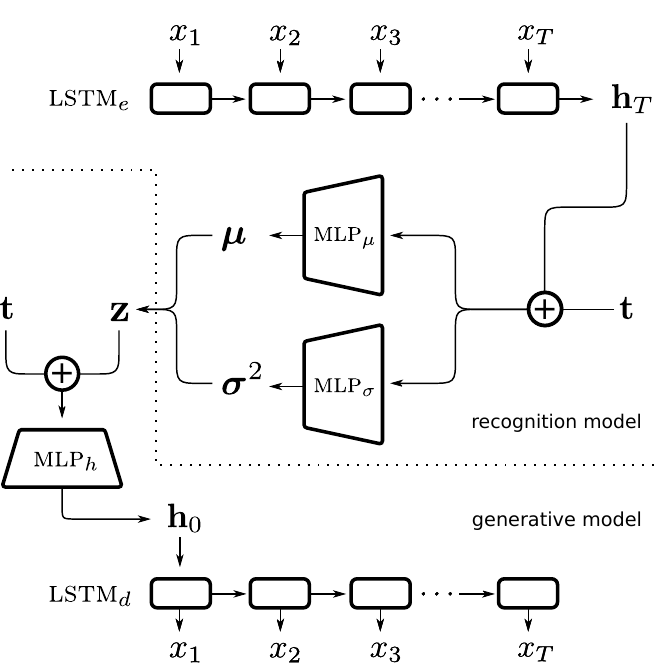}
\\
(a)
\end{minipage}
\hfill
\begin{minipage}[b]{0.45\textwidth}
\centering
\includegraphics[width=2.6in]{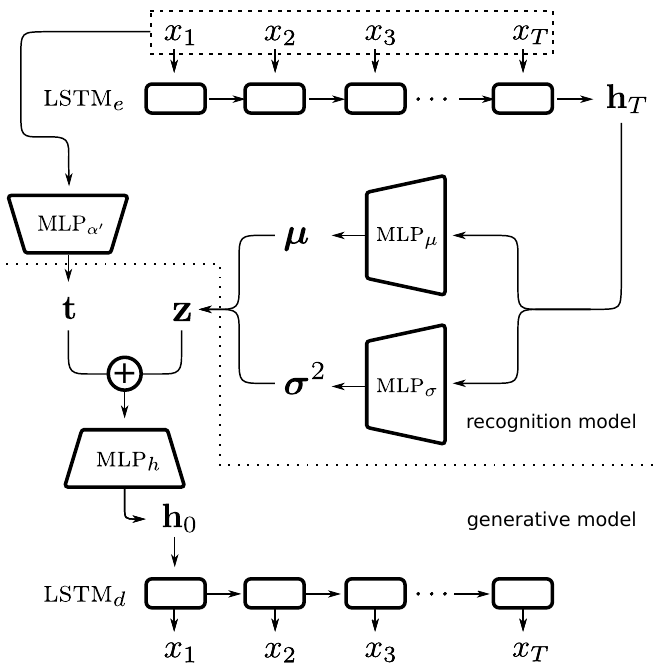}
\\
(b)
\end{minipage}
\caption{Neural network structure for (a) the joint model and (b) the marginal model during training. Area above the dotted line is the recognition model and below is the generative model. Circle with plus sign denotes concatenation operation.}\label{fig:model_structure}
\end{figure*}

\subsection{Proposed Model}
Our formulation of the model involves three random variables: the sentence/document $\vx$, the latent code $\vz$, and a newly introduced topic distribution variable $\vt$. $p(\vz)$, the prior of $\vz$, is a standard multivariate Gaussian with mean $0$ and identity covariance matrix; $p(\vt|\vz)$ is the conditional distributions of $\vt$ given the latent code $\vz$ which follows a Dirichlet distribution. Note here we make the natural choice to model $\vt$ as a dependent variable on $\vz$ because we perceive topics in a sentence as dependent on the abstract meaning. Finally, $p(\vx|\vz,\vt)$ is the text generator. Intuitively, the model mimics a process of generating a sentence: abstract meaning is firstly determined with $\vz$; then a distribution of topics $\vt$ is used depending on $\vz$; finally the sentence $\vx$ is generated by selecting words and constructing a sentence based on both $\vz$ and $\vt$. The base generative model is illustrated in Figure \ref{fig:models} (a).

There are two options when modeling the inference network. As we could obtain topic distributions for all documents using unsupervised topic modeling algorithms such as the latent Dirichlet allocation (\lda) \cite{blei2003latent}, the first option is to model the joint probability of $p(\vx,\vt)$. For the second option, we could incorporate the topic modeling component into the model, treat topic latent as unseen for all training examples, and model $p(\vx)$ directly. We name these two options joint model and marginal model respectively. The recognition models for both the joint and the marginal model are depicted in Figure \ref{fig:models} (b), (c). Note that they share the same generative model illustrated in Figure \ref{fig:models} (a).

\paragraph{Joint model}

Joint model treats $\vt$ as an observed variable. Therefore we need to maximize the joint probability of an input text $\vx$ and its corresponding topic distribution $\vt$. The variational lower bound for the joint model is:
\begin{align}
\label{equ:jointvlb}
\log{p(\vx,\vt)}\geq & \vlb_{\text{J}}(\vtheta,\vphi;\vx, \vt)\nonumber\\
=&\E_{q(\vz|\vx,\vt)}\left[\log{p(\vx|\vz,\vt)}+\log{p(\vt|\vz)}\right]\nonumber\\
&-\kld(q(\vz|\vx,\vt)||p(\vz))
\end{align}

\paragraph{Marginal model}
We could also choose to treat $\vt$ as unobserved and directly model the marginal distribution of $p(\vx)$.
We can derive the \elbo{} of the marginal distribution similarly to Equation \ref{equ:vlb}:
\begin{align}
\label{equ:marginalvlb}
\log{p(\vx)}\geq & \vlb_{\text{M}}(\vtheta,\vphi;\vx)\nonumber\\
= &\E_{q(\vz,\vt|\vx)}\left[\log{p(\vx|\vz,\vt)}\right] \nonumber\\
&-\kld(q(\vz,\vt|\vx)||p(\vz,\vt)) \nonumber\\
=&\E_{q(\vz|\vx)}\left[\E_{q(\vt|\vx,\vz)}[\log{p(\vx|\vz,\vt)}]\right]\nonumber\\
&-\E_{q(\vz|\vx)}\left[\kld(q(\vt|\vx,\vz)||p(\vt|\vz))\right] \nonumber\\
&-\text{D}_{\kl}(q(\vz|\vx)||p(\vz))
\end{align}

During training, the loss function is defined as the negative \elbo{}. Reparameterizable samples from $q(\vz|\vx,\vt)$ (the joint model) or $q(\vz,\vt|\vx)$ (the marginal model) are used to approximate the expectation terms, and \kl{} divergence terms are analytically computed. The gradients could be calculated and the objective can then be optimized with gradient-based optimization methods.

\subsection{Model Specification}
Our choice for the shared generative model is specified as follows:
\begin{align}
p(\vz)&\sim\N(\vzero,\eye) \nonumber\\
\log(\valpha)&=\mlp_\alpha(\vz)\nonumber\\
p(\vt|\vz)&\sim\dir(\valpha) \nonumber\\
p(\vx|\vz,\vt)&=\lstm_d(\mlp_h([\vz,\vt]))\nonumber
\end{align}
where $\mlp_\alpha$ is a multi-layer perceptron that takes the latent code $\vz$ and produces the logarithm of the $\valpha$ to parameterize the Dirichlet prior distribution; $[\vz, \vt]$ denotes the concatenation of the two variables $\vz$ and $\vt$; $\mlp_h$ takes the concatenation and generates the initial hidden state for the decoder $\lstm_d$. $\vx$ is then generated sequentially with $\lstm_d$.

The specifications for the joint recognition model are:
\begin{align}
\vh&=\lstm_e(\vx)\nonumber\\
\vmu, \log(\vsigma^2)&=\mlp_\mu([\vh,\vt]),\mlp_\sigma([\vh,\vt])\nonumber\\
q(\vz|\vx,\vt)&\sim\N(\vmu,\diag(\vsigma^2)) \nonumber
\end{align}
where the input $\vx$ is fed into the encoder $\lstm{}_e$; the last hidden state $\vh$ is then concatenated with $\vt$ and jointly determines the mean and log covariance matrix of the posterior distribution for $\vz$ with $\mlp_\mu$ and $\mlp_\sigma$. The neural network structure for the joint model in the training phase is illustrated in Figure \ref{fig:model_structure} (a).

Similarly, the specifications for the marginal recognition model are:
\begin{align}
\vh&=\lstm_e(\vx)\nonumber\\
\vmu, \log(\vsigma^2)&=\mlp_\mu(\vh),\mlp_\sigma(\vh)\nonumber\\
q(\vz|\vx)&\sim\N(\vmu,\diag(\vsigma^2)) \nonumber\\
\log{(\valpha')}&=\mlp_{\alpha'}(\vx) \nonumber\\
q(\vt|\vx,\vz)&\sim\dir(\valpha') \nonumber
\end{align}

where $\mlp_{\alpha'}$ takes the input $\vx$ and calculates the logarithm of the $\valpha'$ parameterizing the posterior distribution for $\vt$.
The differences from the joint model include: 1. Posterior distribution of $\vz$ does not depend on $\vt$, therefore $\mlp_\mu$ and $\mlp_\sigma$ takes input of only the last hidden state of the encoder $\lstm{}_e$; 2. Posterior distribution of $\vt$ is approximated by making inference given the input $\vx$. Because of the fact that posterior of $\vz$ does not depend on $\vt$, a benefit of the marginal model over the joint model is that we do not need a separate topic modeling component to calculate the topic distributions for the input. The marginal model is also capable of making inference of the topic $\vt$ given input $\vx$. The neural network structure for the marginal model in the training phase is illustrated in Figure \ref{fig:model_structure} (b).

The $\E_{q(\vt|\vx,\vz)}[\log{p(\vx|\vz,\vt)}]$ term in the variational lower bound of the marginal model requires us to sample from a Dirichlet distribution. There are some recent efforts to make the reparameterization trick viable 
\cite{ruiz2016generalized,srivastava2017autoencoding,figurnov2018implicit} for Dirichlet distribution. We follow the method introduced in \cite{jankowiak2018pathwise}.


\begin{table*}[t!]
\centering
\begin{tabular}{lllllllll}
\hline
 \multirow{2}{*}{\bf Model} & \multicolumn{2}{c}{\textbf{\ptb}} & \multicolumn{2}{c}{\textbf{\bc}} & \multicolumn{2}{c}{\textbf{\yahoo}} & \multicolumn{2}{c}{\textbf{\yelp}} \\
 & \nll{} (\kl) & \ppl & \nll{} (\kl) & \ppl & \nll{} (\kl) & \ppl & \nll{} (\kl) & \ppl\\
  \hline
\lstm-\lm & 116.2 & 104.2 & 37.4 & 64.0 & 334.9 & 66.2 & 362.7 & 42.6\\
\vae   &  96.0 (9.6) & 79.8 & 28.0 (6.9) & 22.5 & 337.2 (0.4) & 65.6 & 369.1 (0.5) & 45.5 \\
\vae{} w/ \bow  & 93.2 (12.9) & 70.2 & 17.9 (17.2) & 7.3 & 310.3 (32.0) & 48.7 & 343.3 (31.5) & 34.9\\
\cnn-\vae & - & - & - & - & 332.1 (10.0) & 63.9 & 359.1 (7.6) & 41.1\\
Ours & \textbf{91.9 (2.7)} & 66.2 & \textbf{23.5 (6.6)} & 13.6 & \textbf{327.8 (2.9)} & 60.6 & \textbf{360.3 (2.5)} & 41.6\\
Ours w/ \bow & 76.9 (23.3) & \textbf{33.4} & 15.8 (16.2) & \textbf{5.8} & 308.5 (31.9)  & \textbf{47.6} & 342.9 (30.5) & \textbf{34.7}\\
\hline
\end{tabular}
\caption{Language modeling results on the test sets. \bow{} represents the auxiliary bag-of-words loss \cite{zhao2017learning}. Lowest combined \nll{}+\kl{} and lowest \ppl{} for each data set are highlighted. 
}
\label{tab:lm}
\end{table*}

\section{Experiments and Results}
We experiment both the joint and marginal models in \vae{} and \cvae{} settings. In the following sections, we describe the data sets, detailed setups, and experimental results. Results for the joint and marginal models are combined due to their similar performances. The advantage of not requiring an external topic model component makes the marginal model a preferable choice in practice.
\subsection{Data Sets}
We want to explore how effective the topic distributions along with the latent variable help to reconstruct inputs and generate new samples. Therefore we choose to adopt four different datasets, two with relatively short sentences (the Penn Treebank (\ptb) \cite{marcus1993building} and the BookCorpus (\bc) \cite{moviebook}), and two with relatively long documents (Yahoo Answer (\yahoo) and Yelp15 review (\yelp)) in the \vae{} setting to see whether the latent representations contribute to a better reconstruction of the inputs. We further extract the latent features of the \yahoo{} and \yelp{} data sets learned by their corresponding \vae{} models and train classifiers with various amount of training data to evaluate the qualities of the learned representations. We also train \cvae{} versions of the proposed model on \yahoo{} and \yelp{} data sets to evaluate the conditional generation capabilities of the proposed model. Notice we do not compare with topic modeling studies because topic modeling component is only present in the marginal model and is not the focus of this study.

For \bc{}, we randomly sample 200k sentences for training, 10k for validation and 10k for test. Subsets of \yahoo{} and \yelp{} used in \cite{yang2017improved} are adopted. Each subset contains 100k documents for training, 10k for validation and 10k for test. \yahoo{} contains ten different classes representing topics for each document; \yelp{} has five levels of ratings for each document. In both data sets, the class labels are balanced.

\subsection{Experiment Setup}
For the inputs, we set a maximum vocabulary size of 20k and a maximum length of 200 across all data sets. We train \lda{} models for each data set using only the training data portion with number of topics in [8, 16, 32, 64]. These \lda{} models are then used to process all splits of data to obtain topic distributions for use in the joint models.

We use one-layer \lstm{} with hidden size [200, 512] for both the encoder and the decoder. Word embeddings are shared and size is set to 200. Embedding size for the class labels is 8 in the \cvae{} setting. We choose the dimension of $\vz$ from [16, 32]. $\mlp_\mu$, $\mlp_\sigma$ and $\mlp_h$ are two-layer fully connected neural networks with hidden size same as the encoder/decoder. $\mlp_{\alpha}$ and $\mlp_{\alpha'}$ are also two-layer \mlp{}s with outputs exponentiated to ensure positivity of the Dirichlet parameters.

All models are trained with Adam \cite{kingma2014adam} with learning rate 1e-3. Weight decay is chosen from [1e-3, 3e-3, 1e-2]. Dropout ratio for both the encoder and the decoder is 0.2. We use batch size of 32 and models are trained for 48 epochs. \kl{} annealing is used with a linear scheduling from 0 at step 2k to 1 at step 42k. No word dropping is applied.

We also conduct experiments of classification with learned representations. We train \vae{} models on both \yahoo{} and \yelp{} data sets and extract latent features for all data splits. We randomly choose 500, 2000 training samples to train a linear support vector machine (\svm{}) on the latent representations and test performances on the test sets. 

\begin{table}[t!]
\begin{center}
\begin{tabular}{lcccc}
\hline
\multirow{2}{*}{\bf Model} & \multicolumn{2}{c}{\bf \yahoo} & \multicolumn{2}{c}{\bf \yelp} \\ 
 & 500 & 2000 & 500 & 2000 \\ \hline
\vae & 51.6 & 53.2 & 25.1 & 27.2 \\
\vae{} w/ \bow & 52.1 & 54.4 & 38.2 & 40.0 \\
Ours & 52.5 & 55.2 & 32.2 & 34.6\\
Ours w/ \bow & \bf 53.3 & \bf 57.6 & \bf 39.8 & \bf 42.4\\ 

\hline
\end{tabular}
\end{center}
\caption{\label{tab:rep_clf}Test set accuracy (\%) of classifiers trained with learned representations from different \vae{} models. Numbers reported are averaged over five experiments.}
\end{table}

\begin{table}[t!]
\begin{center}
\begin{tabular}{lcccc}
\hline
\multirow{2}{*}{\bf Model} & \multicolumn{2}{c}{\bf \yahoo} & \multicolumn{2}{c}{\bf \yelp} \\ 
 & \cnn & \lr & \cnn & \lr \\ \hline
 \textsc{std samps.} \\
\cvae & 15.8 & 14.5 & 30.4 & 24.9 \\
\cvae{} w/ \bow & \bf 50.9 & 58.5 & \bf 44.7 & 45.9 \\
Ours & 37.0 & 50.6 & 37.6 & 38.9\\
Ours w/ \bow & 50.0 & \bf 59.0 & 43.2 & \bf 46.0\\ \hline
 \textsc{lp samps.} \\
\cvae & 20.8 & 18.7 & 32.2 & 29.2 \\
\cvae{} w/ \bow & 37.8 & \bf 54.8 & 21.2 & 20.4 \\
Ours & \bf 38.7 & 51.5 & 37.4 & 39.2\\
Ours w/ \bow & 33.8 & 52.7 & \bf 42.0 & \bf 46.3\\

\hline
\end{tabular}
\end{center}
\caption{\label{tab:clf}Test set accuracy (\%) of classifiers trained with generated data from different \cvae{} models. \textsc{lp samps.} stands for low-probability samples. Numbers reported are averaged over five experiments.}
\end{table}

\begin{table*}[t!]
\small
\begin{center}
\begin{tabular}{cll}
\hline
\textsc{input} & \bf Tears filled his eyes. & \bf He takes a sharp breath, and opens his eyes.\\
\textsc{mean} & Tears welled his eyes. & He got a seat and reached for the door.\\ \hline
$\vz$ \textsc{samp}. 1 & \it Sofia rubbed her eyes. & \it He pulled his head and continued to stare at me.\\
$\vz$ \textsc{samp}. 2 & \it His eyes widened in anticipation. & \it The captain scowled and took a seat in the courtyard.\\
$\vz$ \textsc{samp}. 3 & \it Haleton ignored his eyes. & \it She pulled a hand on the ground and turned.\\ \hline
$\vt$ \textsc{samp}. 1 & \it Jason asks his voice. & \it He looked at the boy, sitting on his shoulder.\\
$\vt$ \textsc{samp}. 2 & \it The young ninja nodded. & \it He asked, a little shocked, raising his eyebrows.\\
$\vt$ \textsc{samp}. 3 & \it Frank shook his head. & \it He offered a seat and pressed the ground in his face.\\
\hline
\end{tabular}
\end{center}
\caption{\label{tab:recs}Examples of generated sentences from the posterior distributions of $\vz$ and $\vt$. $\vt$ is fixed to the mean when sampling from $\vz$ and vice versa. We can observe that $\vt$ to some extent controls the word choices and topics while $\vz$ relates to the sentence structures.} 
\end{table*}

\subsection{Language Modeling Results}
The language modeling results are shown in Table \ref{tab:lm}. We report the negative log-likelihood (\nll), the \kl{} divergence term between the posterior and the prior of $\vz$, as well as the perplexity (\ppl).

We compare with \lstm{} language model, standard \vae{}, standard \vae{} with bag-of-word loss (\bow) \cite{zhao2017learning}, and dilated convolutional \vae{} \cite{yang2017improved}. \textsc{di}-\vae{} \cite{zhao2018unsupervised} uses multi-way categorical instead of multivariate Gaussian for the latent variable, therefore the results are not directly comparable. We also want to stress that \ppl{} only measures one aspect of the model and should not be considered the only metric for this task. Our model achieves best combined results (\nll+\kl) across all data sets. We observe from Table \ref{tab:lm} that when trained with \bow{} loss, \nll{} is noticeably reduced while the \kl{} divergence is much larger. The combined loss \nll+\kl{} is always larger compared to model trained without \bow{}. Compared to the baseline \vae{} model, the proposed model achieves better \nll{} and \kl{} values at the same time. One possible reason is the multi-modality of the Dirichlet variable $\vt$ help ease the labor for $\vz$ which is a unimodal distribution. 

It is worth noting that for \yahoo{} and \yelp{} data sets, the original \vae{} fails to learn latents with reasonably large \kl{} divergence (0.4 for \yahoo{} and 0.5 for \yelp) even with the \kl{} annealing trick. This, we conjecture, could be due to the relatively long document lengths in the two data sets. It is more rewarding for the model to learn a powerful decoder to maximize the overall data probability. Whereas our model learns latents with non-trivial \kl{} divergence across all data sets.

\subsection{Classification on Learned Representations}
For \yahoo{} and \yelp{} data sets, we train both \vae{} without the label information and \cvae{} with label information. We use the trained \vae{} models to extract latent representations for all the data splits and randomly choose 500 and 2000 examples from the training split to train a linear \svm{} on the feature space. Model selections are done with a validation set and test performances are shown in Table \ref{tab:rep_clf}. Evaluation of the trained \cvae{} models are described in the next section.

From the results we can tell that using the latent representations learned with topic-aware \vae{} models, we get better classification performances compared to standard \vae{} models. This indicates that the proposed models learn better representations when used for classification. We can also see from Table \ref{tab:rep_clf} that models trained with \bow{} loss extract better features.

\subsection{Conditional Generation Results}
We adopt an evaluation approach for \cvae{} detailed as follows: 
1. Generate random samples conditioned on different label classes with the trained model;
2. Train a classifier with the generated data to predict the document labels. Select the best classifier based on performances on the original validation set; 
3. Evaluate the classifier on the original test set.

This evaluation protocol considers both the reconstruction quality and the learned latent space. If the reconstructions are not good enough, the generated examples would be too deviated from the true data distribution and the supervision they provide to the classifier is unreliable. On the other hand, if the model memorize perfectly each example by encoding them into disjoint points in the latent space, random samples from this latent space are highly unlikely to fall into non-zero density areas (for a discussion on the \nll{}/\kl{} trade-off, see the model analysis section).

We use both a \cnn{} classifier described in \cite{kim2014convolutional} and a logistic regression model with tf-idf features. Same number of examples as the original training set are generated with each \cvae{} model and all experiments are repeated five times to account for randomness during sample generation and classifier training. We include results for standard samples and low-probability samples. Low-probability samples are generated by scaling the standard samples by 2 from the corresponding mean value. The results are shown in Table \ref{tab:clf}. Models trained with auxiliary \bow{} loss are favorable by this evaluation protocol with their standard samples. When generating from the low-probability areas, however, \cvae{} with \bow{} degrades sharply while the proposed model is able to maintain similar if not better samples.

\begin{table}[t!]
\small
\begin{center}
\begin{tabular}{cl}
\hline
\multirow{5}{*}{both} & \bf He looked at the small buildings and smiled.\\ 
 &  \it I looked at the picture, confused. \\
 &  \it I frown at the young man. \\
 &  \it I knew what the man was hearing. \\
 & \bf I knew what was going on. \\ \hline
\multirow{5}{*}{$\vz$} & \bf He looked at the small buildings and smiled. \\
& \it I looked at the bar, stunned. \\
& \it I looked at the bar, stunned. \\
& \it I thought he was a bad friend. \\
& \it I thought he was a teenager. \\ \hline
\multirow{5}{*}{$\vt$} & \small \bf He looked at the small buildings and smiled. \\
& \it He looked at the young man and listened intently. \\
& \it He straightened his face and saw the agents. \\
& \it He straightened his face and saw the voice. \\
& \it He glanced at the young man and stood.\\
\hline
\end{tabular}
\end{center}
\caption{\label{tab:ints}Generated sentences from interpolations of points in the latent space. Three interpolation types are: interpolate both $\vz$ and $\vt$ (top), interpolate $\vz$ only (middle), and interpolate $\vt$ only (bottom). Notice how sentence structure barely varies when $\vz$ is kept fixed and how varying $\vt$ affects the word choices.}
\end{table}

\section{Model Analysis}
In this section, we qualitatively evaluate properties of the model by analyzing the generated samples in different scenarios. We also include a discussion on the trade-off between \kl{} and \nll{} terms in the training objective.

\subsection{Reconstruction}

The model is able to reconstruct input documents by first encode them into latent space and then decode from the mean or samples of the posterior distribution of $\vz$. For the marginal model, we could also sample from the $\vt$ space. Table \ref{tab:recs} shows two examples of generated sentences with a greedy decoder from means and samples of the posteriors of $\vz$ and $\vt$. The model does not memorize exactly the inputs as we can see from the examples. Instead, sentences with similar structures and tenses are generated. Notably, in the first example, all three $\vz$ samples preserve the topic word ``eyes'' because they share the same $\vt$ value.

\subsection{Interpolation}

One property of \vae{} models is that the models are able to generate smoothly transitioned sentences from interpolations of points in the latent space \cite{bowman2015generating}. For our proposed model, interpolations can be done on both $\vz$ and $\vt$. Table \ref{tab:ints} shows examples of interpolation of all three types. We could observe from the examples that the model is able to generate smoothly transitioned sentences in both $\vz$ and $\vt$ spaces. More specifically, $\vt$ space is somehow independent of the general sentence structure: all five sentences have the same pronoun and similar structures. This is true for most of the interpolations generated. 
While the effect of $\vt$ seems to be determining the general topics and choice of words in the sentences, the effect of $\vz$ is harder to isolate and is possibly entangled with the effect of $\vt$.

\subsection{Topics Generated by the Marginal Model}

As mentioned in previous sections, the marginal model is able to generate topic distributions based on the input text at test time. We examine the topic distributions generated from a marginal model trained on the \yelp{} data set with 8 topics. Top words for each topic are listed in Table \ref{tab:topic_words}. Notably, topic 1 contains words for Japanese restaurants; topic 3 contains words used by non-English reviewers; topic 4 is about hotels; topic 5 relates to descriptions of service time; topic 8 is about breakfast.

\begin{figure}\centering    
	\includegraphics[width=3.in]{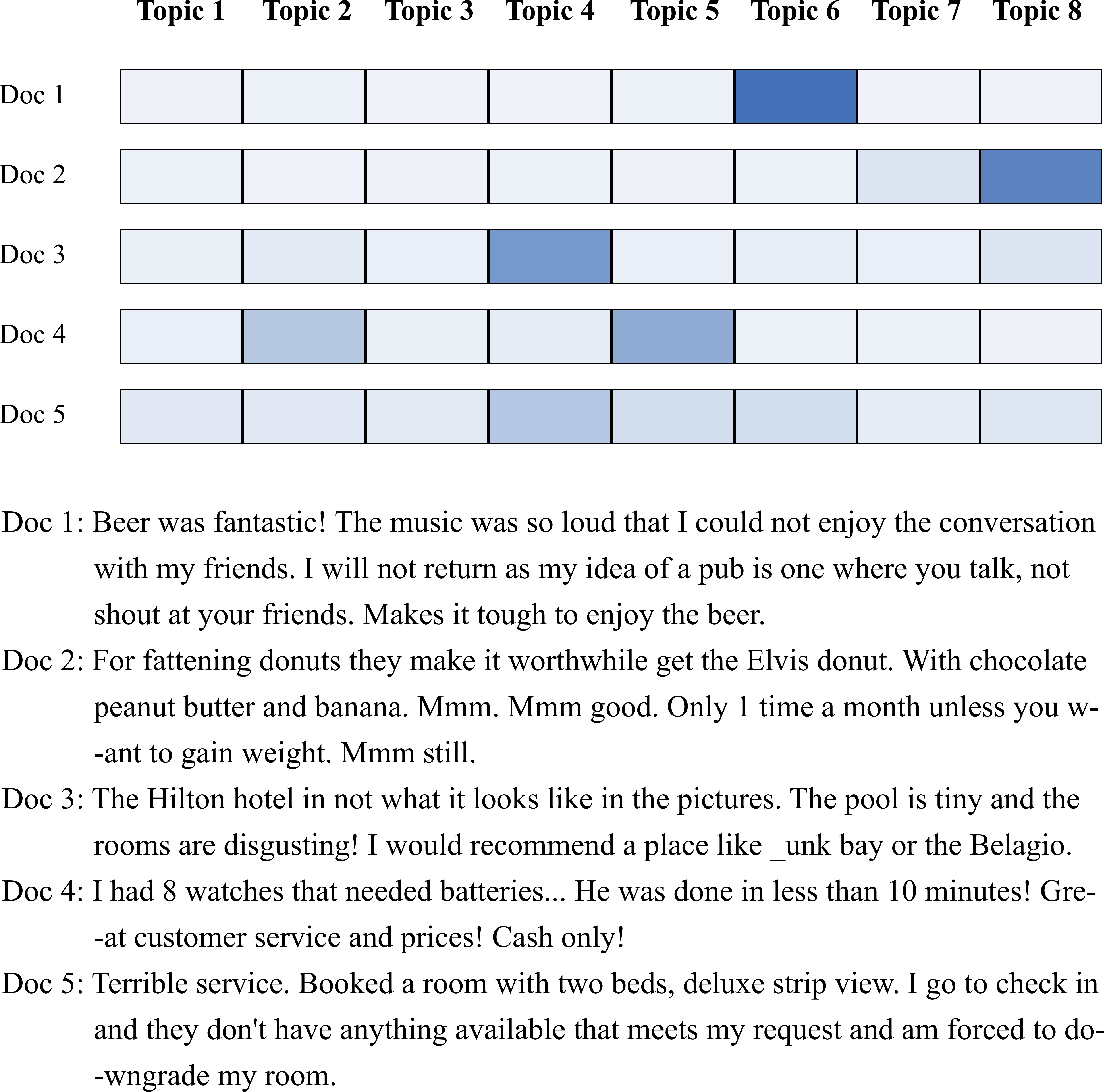}
	\caption{Examples of generated topic distributions. Color changes linearly with the probability density. Darker means higher probability.}\label{fig:topics}
\end{figure}

Five examples of the generated topic distributions are shown in Figure \ref{fig:topics}. We can see that the generated topic distributions in most cases are representative of the reviews. There are other documents with more uniformly distributed topics. We want to stress here that the proposed model is not designed to generate high-quality topic distributions. This functionality is a side-product of the proposed marginal model.

\begin{table}
\small
\begin{center}
\begin{tabular}{cl}
\hline
\bf Topic & \bf Top words \\ \hline
1 & sushi, fast, food, rolls, roll, hair, egg \\
2 & place, like, great, just, time, ve \\
3 & la, taking, die, le, dr, et, son, est \\
4 & room, hotel, vegas, stay, strip, guys\\
5 & did, service, minutes, got, time, said\\
6 & food, great, place, good, service, bar\\
7 & good, food, place, great, chicken\\
8 & like, breakfast, coffee, good, buffet \\
\hline
\end{tabular}
\end{center}
\caption{\label{tab:topic_words} Top words for each topic from an \lda{} model trained on the \yelp{} data with 8 topics.}
\end{table}

\subsection{Trade-offs Between \kl{} and \nll{}}

By the nature of the \vae{} model, there is a trade-off between the \nll{}, which measures the reconstruction quality, and the \kl{} term, which prevent the model from simply memorizing training samples. We conduct a experiment on a synthetic data set to better understand the behavior of \vae{} when \kl{} divergence is at different levels.

\begin{figure}\centering    
	\includegraphics[width=3.03in]{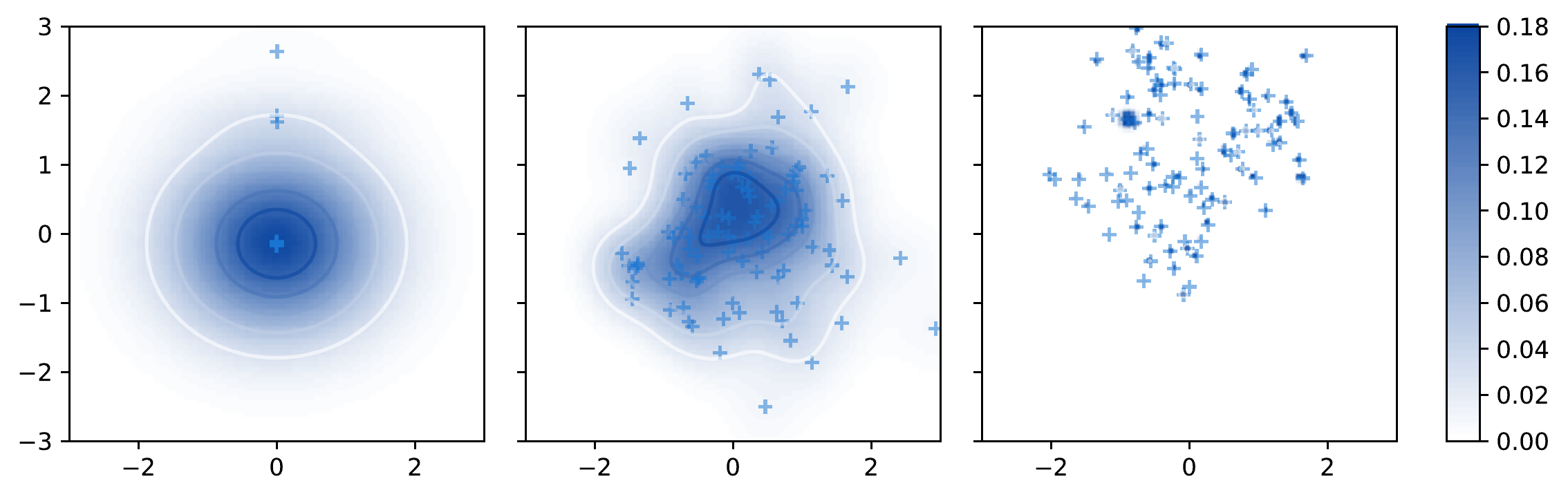}
	\caption{\label{fig:kl_comp}Averaged latent code $\vz$ distributions for \textsc{sm}-\kl{} (left), \textsc{md}-\kl{} (middle), and \textsc{lg}-\kl{} (right) representing models with different scales of \kl{} values. Plus symbols are latent variable means. When \kl{} value is too small, learned representations are random; when \kl{} is too large, the model encodes each training example into a single point essentially memorizing the input.}
\end{figure}

The synthetic data set is constructed with 100 unique tokens. Token transition probabilities are randomly initialized. 12k examples are sampled with maximum length 8. 10k are used as training, 1k for validation and 1k for test. We set the embedding size and hidden size as 16 and the code size as 2. Three models are trained with different levels of weights on the \kl{} divergence term. The \nll{} (\kl) performances are 34.0 (0.2), 32.2 (1.9), and 28.2 (9.9). We term these three models \textsc{sm}-\kl, \textsc{md}-\kl, and \textsc{lg}-\kl. Distributions of the latent code $\vz$ for 100 randomly selected test samples are averaged and shown in Figure \ref{fig:kl_comp}.

We can see that when \kl{} term is very small, almost all samples get encoded into the same uniform Gaussian. When \kl{} term is at a medium level, encoded samples are more scattered but still have non-zero densities when transitioning from the mean value of a sample to another. When \kl{} term is very large, the mean values of the encoded samples are still not far away from each other in the latent space. What is different is that samples are encoded into disjoint areas with very small covariances. This indicates that the model achieves higher reconstruction qualities by trying to memorize each sample. Figure \ref{fig:kl_comp} indicates that there is an optimal level of \kl{} divergence for \vae{} models. However, how to determine this optimal level of \kl{} divergence remains an open problem.

\section{Conclusion}
We present a novel variational autoencoder with a Dirichlet latent variable representing text topic distributions. Learning objectives of the proposed model in two different settings are derived and experiments on four different data sets are conducted. The model is more robust against \kl{} divergence vanishing and learns better latent representations compared to a standard \vae{} model with respect to classification performances. It also generates higher quality samples across the latent space, even from low-probability areas. Interpolations of points in the latent space can be used to generate plausible sentences with smoothly transitioned semantics. As a side product of the proposed formulation, the model can generate topic distributions of unseen documents. We discuss the \kl{}/\nll{} loss term trade-off and experiment on a toy data set which shows that there is an appropriate level of \kl{} divergence depending on the application and model settings. Our findings also suggest some promising future research directions, including systematically determination of the proper \kl{} divergence level based on model settings, learn disentangled representations of $\vz$ and $\vt$.

\bibliography{naaclhlt2019}
\bibliographystyle{acl_natbib}
\end{document}